# Chapter 1
# Energy-Efficient Autonomous Driving Using Cognitive Driver Behavioral Models and Reinforcement Learning

Huayi Li, Nan Li, Ilya Kolmanovsky, and Anouck Girard

**Abstract** Autonomous driving technologies are expected to not only improve mobility and road safety but also bring energy efficiency benefits. In the foreseeable future, autonomous vehicles (AVs) will operate on roads shared with human-driven vehicles. To maintain safety and liveness while simultaneously minimizing energy consumption, the AV planning and decision-making process should account for interactions between the autonomous ego vehicle and surrounding human-driven vehicles. In this chapter, we describe a framework for developing energy-efficient autonomous driving policies on shared roads by exploiting human-driver behavior modeling based on cognitive hierarchy theory and reinforcement learning.

## 1.1 Introduction

With recent advances in sensing technologies and artificial intelligence, there has been a rapidly growing interest in connected and autonomous vehicles (CAVs) [1, 2]. Such vehicles are expected to improve the safety and mobility of transportation and to alleviate traffic congestion.

Another expected benefit of CAVs is a reduction in fuel/energy consumption [3, 4]. Since 2016, the U.S. Department of Energy has awarded more than $50 million in funding for studies by the Advanced Research Projects Agency-Energy's (ARPA-E) Next-Generation Energy Technologies for Connected and Automated On-Road Vehicles (NEXTCAR) program for which the goal is to reduce the energy consumption of vehicles in all classes by more than 20% via CAV technologies

This research was supported by Mcity, University of Michigan.
This research was also supported in part through computational resources and services provided by Advanced Research Computing at the University of Michigan, Ann Arbor.
The authors are with the Department of Aerospace Engineering, University of Michigan, Ann Arbor, MI 48109, USA (e-mail: huayil@umich.edu; nanli@umich.edu; ilya@umich.edu; anouck@umich.edu).

[5]. Table 1.1 shows selected results by this program. This is not intended to be a comprehensive review, but the collection illustrates the variety of methods and traffic conditions being explored in some of the most recent studies to achieve energy efficiency improvements using CAV technologies.

**Table 1.1** Summary of selected literature on using CAV technologies to improve energy efficiency

| Ref. | Energy consumption reduction by [%] | Methods | Traffic conditions |
|---|---|---|---|
| [6] | 5.4 | Thermal management (air conditioning) | Drive cycles |
| [7] | 3.1 | Thermal management (battery) | Drive cycles |
| [8] | 41 | Thermal management (air conditioning and engine) | Drive cycles |
| [9] | 50 | Eco-driving | Intersections |
| [10] | 12 | Eco-routing | City |
| [11] | 57.8 | Platooning, cooperative merging | Ramps on highway |
| [12] | 8.5 | Anticipative lane change | City and highway |
| [13] | (up to) 59 | Cooperative lane change | Highway |
| [14] | 47 | Cooperative driving | Intersections and roundabouts |

To realize this encouraging potential in real-world driving circumstances, we observe that at least the following two problems remain to be studied. Firstly, interactions of the ego vehicle with surrounding vehicles in traffic need to be considered. In the foreseeable future, autonomous vehicles will operate together with human-driven vehicles in traffic. Thus, it is necessary to consider the different vehicle actions and reactions caused by different types of human driving styles. In [15, 16, 17, 18], level-$k$ game theory is used to model the interactions with the focus on different driving scenarios. Researchers such as of [19, 20, 21] have utilized traffic-in-the-loop models and closed-loop control to achieve simultaneous optimization for safety and fuel economy. However, only longitudinal control is considered in these studies. Indeed, the vast majority of recent studies on improving energy efficiency using CAV technologies assume that the ego vehicle is driven in single-lane traffic. Thus, the second problem worth investigating is the simultaneous longitudinal control and lateral control (such as lane changes) of AVs, which increases the dimension of the problem but provides additional possibilities to save energy. A more detailed discussion on lane changes for better energy efficiency is given in Section 1.2.

There has been a rich set of research on machine learning (ML) methods for automotive applications to improve energy efficiency and emissions by modeling and control of the powertrain system (see, e.g., [22, 23, 24, 25, 26, 27]). In particular, to meet increasingly stringent fuel economy and emissions regulations, the powertrain systems of hybrid electric vehicles (HEVs) have become more and more complex. Consequently, commonly studied model-based control methods for the energy management system (EMS), such as dynamic programming (DP) and model predictive control (MPC) [28], are facing growing difficulties, as they rely on models with good accuracy and many control-oriented models are physics-based. In comparison, machine learning methods can handle this challenge well. For example, for HEVs with small electrical energy storage, there is a significant potential to utilize recurrent neural networks to learn driving patterns and improve energy efficiency [29]. In the CAV domain, example applications of ML include perception and localization, route/path planning/optimization, and motion control, despite challenges such as computation, safety, and adaptability/generalizability that are actively being studied [30, 31, 32]. Studies such as [33, 34] use ML to inform energy-efficient acceleration/braking of electric vehicles. The authors previously developed a level-$k$ game theory-based traffic simulator in [15] (following the methodology originally proposed in [35]). The simulator is based on cognitive driver behavioral models trained by reinforcement learning (RL).

In this chapter, we describe a novel framework for developing energy-efficient AV control policies (including both longitudinal (speed) and lateral (lane change) controls) through RL. We focus our attention on highway driving and autonomous battery electric vehicles (BEVs), as BEVs are getting increasingly popular due to their environmental benefits [36]. A BEV powertrain model is developed to calculate the energy consumption over trips. To enable the AV control policy to properly respond to the interactions with human-driven vehicles on shared roads, the game-theoretic traffic model developed in [15, 16] is used as the RL training environment. Reference [37] is a preliminary conference version of this chapter. Extensions to other powertrain types and traffic environments are possible [38, 18] but are left to future work.

The remainder of this chapter is organized as follows. Firstly, further background on lane changes for energy-efficient AV driving is discussed in Section 1.2. Then, we begin the development by building a BEV model and validating it in Section 1.3. In Section 1.4, the control development is detailed, and another control policy trained by RL and the finite-state-machine (FSM) controller from [16] are introduced to be subsequently used for comparison. Section 1.5 presents results on RL convergence and on performance of the developed control policy in simulations. Finally, concluding remarks are made in Section 1.6.

## 1.2 Lane Changes for Energy-Efficient AV Driving

Including lateral actions such as lane changes may further improve the energy efficiency [39], though the survey results of [3] show that this is still an emerging area that remains to be studied. Anticipating lane selection has been proposed, such as in [40, 41]. Furthermore, instead of focusing on an individual vehicle, cooperative lane change [42] is expected to benefit the neighboring vehicles and harmonize the surrounding traffic. However, these studies assume having connected vehicle technologies, such as vehicle-to-vehicle and vehicle-to-infrastructure communications. A general observation is that the energy efficiency can be improved by reducing the change of speed and acceleration. In contrast, our work uses the position and velocity information of the immediate neighboring vehicles detected by the sensors of the ego vehicle, and a powertrain model is used to accurately predict the energy consumption.

There are two major challenges to combine longitudinal and lateral actions for safe energy-efficient driving. Firstly, considering both longitudinal and lateral actions increases the problem complexity. In particular, there are subtle trade-offs between safety and energy efficiency. For instance, in scenarios such as a sudden cut-in by a slow vehicle, changing lanes rather than hard braking preserves vehicle momentum and avoids energy loss, but if the traffic density is high, performing a lane change may not be safe or feasible.

Another challenge is that, unlike safety constraint violation scenarios (e.g., collisions) where events typically occur within seconds, energy efficiency evaluation requires longer time horizons of several hundreds of seconds. Hence, optimization-based control algorithms that simultaneously address driving safety and energy efficiency requirements need to account for both short-term and long-term objectives. One approach is to define a terminal cost function for the short horizon optimization reflective of long-term rewards. However, how to practically determine such a terminal cost function is often *a priori* unclear. In special cases, the problem can be reformulated to focus on maintaining component operation in more efficient regions [43] for which short-horizon optimization is sufficient. However, such reformulations are not always feasible and the performance with such an approach could be sub-optimal. In particular, Stackelberg policies and the decision tree policies considered in [44, 45, 46] rely on rewards being evaluated short-term. It may not be straightforward to extend these to account for energy efficiency requirements. For our AV control policy, since the RL algorithm updates the value functions with not only the one-step/instantaneous reward but also the average reward over time, it is able to handle multiple optimization objectives that need to be evaluated over different time horizons.

## 1.3 Powertrain Modeling for Battery Electric Vehicles

### 1.3.1 Model Description

Accounting for energy efficiency in the controller design requires a longitudinal powertrain model. As an RL process is generally computationally intensive, a powertrain model used in RL should have low computational footprint but sufficiently high accuracy.

Fig. 1.1 shows the layout of the BEV model considered in this work. The powertrain system consists of a battery pack, a motor/generator (MG), a drivetrain with a single-speed final drive, the wheels and tires, and the powertrain control unit (PCU).

The powertrain model is of the backward type [28, 47, 48]. (See Remark 1 below for details.) It has two states, state of charge ($SOC$) and total energy consumption. High-fidelity dynamics, such as transient responses of the MG and drivetrain, are not considered. Maps (i.e., look-up tables) are used to represent component operating characteristics as described below. A benefit of using a map-based approach is that it can facilitate the change of component specifications so that potential extensions, such as component sizing or fleet energy efficiency studies, are possible.

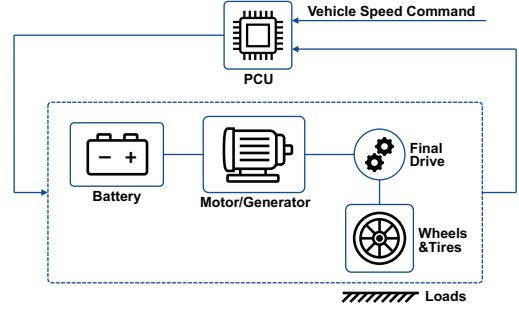

**Fig. 1.1** Electric vehicle powertrain system layout.

*Remark 1* A backward powertrain model assumes that the actual vehicle speed is always equal to the reference speed command. Rotational speeds of components are coupled/scaled by the gear ratios and wheel radius based on this command. The torque required at the wheels to meet the acceleration demand is first calculated and then translated component by component to the actuators such as the motor and the brakes. Backward models typically entail low computational costs and are suitable for (rough) energy consumption evaluations. In comparison, forward models involve a driver model, typically modeled as a PID controller, that commands the motor/brake torque in order to track the vehicle speed reference. The torque is then translated to the wheels through drivetrain components. The speed of each component, as well as the actual vehicle speed, is calculated by integrating the acceleration produced by the torque. Consequently, forward models can better capture the component dynamics, at the cost of higher computational complexity than backward models. They are typically used for more detailed (e.g., componentwise) energy efficiency analysis as well as drivability-related simulations [28, 47, 48]. Also, the forward model is not as suitable as the backward model in our case considering safe driving since there is tracking error between the commanded and actual vehicle speed whose value depends on the tuning of the control parameters.

In the BEV model, the MG speed and traction torque at the wheels are first calculated based on the vehicle speed command according to

$$\omega_{mg} = \frac{V \cdot g}{r}, \tag{1.1}$$

$$T_a = \dot{V} \cdot M \cdot r, \tag{1.2}$$

$$T_l = a + b \cdot V + c \cdot V^2, \tag{1.3}$$

$$T_{whl} = T_a + T_l, \tag{1.4}$$

where $\omega_{mg}$ is the MG speed, $V$ is the vehicle speed, $g$ is the final drive gear ratio, $r$ is the effective wheel radius, $T_a$ is the acceleration torque, $M$ is the effective vehicle mass involving powertrain component inertia, $T_l$ represents lumped external loads including rolling and aerodynamic resistance (while the grade is assumed to be zero) approximated by a quadratic function with coefficients $a$, $b$, and $c$, and $T_{whl}$ is the traction torque at wheels.

The PCU then checks whether the traction torque demand exceeds the battery or MG limits and distributes the torque command to the MG and the friction brake, using the following logic,

$$T_{mgPos} = \begin{cases} 0, & \text{if } T_{whl} < 0, \\ T_{whl} \cdot \frac{1}{g}, & \text{if } 0 \leq T_{whl} \leq T_{mgMax} \cdot g, \\ T_{mgMax}, & \text{if } T_{whl} > T_{mgMax} \cdot g, \end{cases} \quad (1.5)$$

$$T_{whlBrk} = \begin{cases} 0, & \text{if } -T_{whl} < 0, \\ -T_{whl}, & \text{if } 0 \leq -T_{whl} \leq T_{brkMax}, \\ T_{brkMax}, & \text{if } -T_{whl} > T_{brkMax}, \end{cases} \quad (1.6)$$

$$T_{mgReg} = \begin{cases} F_{reg} \cdot T_{whlBrk} \cdot \frac{1}{g}, & \text{if } T_{whlBrk} \leq T_{mgRegLim} \cdot g, \\ F_{reg} \cdot T_{mgRegLim}, & \text{if } T_{whlBrk} > T_{mgRegLim} \cdot g, \end{cases} \quad (1.7)$$

$$T_{mg} = \begin{cases} T_{mgPos}, & \text{if } T_{whl} > 0, \\ -T_{mgReg}, & \text{if } T_{whl} \leq 0, \end{cases} \quad (1.8)$$

$$T_{mechBrk} = -(T_{whlBrk} - T_{mgReg}), \quad (1.9)$$

where $T_{mgMax}$ is the maximum MG torque limit, $T_{mgPos}$ is the positive portion of the MG torque limited by $T_{mgMax}$, $T_{whlBrk}$ is the negative portion of the traction torque at wheels limited by a constant brake torque limit denoted by $T_{brkMax}$, $T_{mgRegLim}$ is the MG regeneration torque limit, $F_{reg}$ is the regeneration factor, $T_{mgReg}$ is the negative portion of the MG torque limited by $T_{mgRegLim}$, $T_{mg}$ is the MG torque, and $T_{mechBrk}$ is the torque demand assigned to the friction brakes. We obtain the values of $T_{mgMax}$, $T_{mgRegLim}$, and $F_{reg}$ through maps, where both $T_{mgMax}$ and $T_{mgRegLim}$ depend on $\omega_{mg}$, and $F_{reg}$ is a function of $V$ (to reduce the regenerative braking at low vehicle speeds) and the battery $SOC$ (to reduce the regenerative braking at high $SOC$).

The power drawn by the MG is then obtained from

$$P_{mg} = \begin{cases} T_{mg}\omega_{mg}/\eta, & \text{if } T_{mg}\omega_{mg} \geq 0, \\ T_{mg}\omega_{mg} \cdot \eta, & \text{if } T_{mg}\omega_{mg} < 0, \end{cases} \quad (1.10)$$

where $P_{mg}$ is the MG electric power, and $\eta \in (0, 1)$ is the MG efficiency as a function of the MG torque and speed, as given by a map.

To calculate the *SOC* and cumulative energy consumption, the battery is modeled as follows:

$$P_{mg} = (V_{oc} - \frac{1}{N_p} \cdot I \cdot R) \cdot N_s \cdot I$$
$$= N_s \cdot V_{oc} \cdot I - \frac{N_s}{N_p} \cdot R \cdot I^2, \quad (1.11)$$

which can be re-arranged as

$$\frac{N_s}{N_p} \cdot R \cdot I^2 - N_s \cdot V_{oc} \cdot I + P_{mg} = 0, \quad (1.12)$$

where $V_{oc}$ and $R$ are, respectively, the open circuit voltage and the resistance of a single battery cell, $I$ is the battery pack current, $N_s$ is the number of battery cells in series, and $N_p$ is the number of battery cells in parallel. Here, $V_{oc}$ and $R$ are acquired through maps, and both variables depend on *SOC*, with the assumption that the battery temperature is constant. Then, we can solve for the battery current as

$$I = \frac{N_s \cdot V_{oc} - \sqrt{(N_s \cdot V_{oc})^2 - 4\frac{N_s}{N_p} \cdot R \cdot P_{mg}}}{2\frac{N_s}{N_p} \cdot R}, \quad (1.13)$$

and the battery dynamics are given by

$$\dot{SOC} = \frac{-I}{C_{max} \cdot N_p \cdot 3600} \cdot 100, \quad (1.14)$$

where $C_{max}$ is the maximum battery capacity.

The total discharged electric energy $E_{batt}$ is computed by integrating the battery power $P_{batt}$ based on

$$\dot{E}_{batt} = P_{batt} = N_s \cdot I \cdot V_{oc}, \quad (1.15)$$

and the energy consumption can be determined from

$$MPGe = \frac{x}{E_{batt}} \cdot \gamma, \quad (1.16)$$

where $MPGe$ stands for miles per gallon equivalent, $x$ is the total distance traveled, and $\gamma$ represents the unit conversion coefficient.

### 1.3.2 Model Calibration and Validation

The BEV powertrain model described above is calibrated using the BEV reference model in the Powertrain Blockset Toolbox (PTBS) version 1.5 developed by MathWorks. The PTBS reference model is a forward model and includes more detailed component controls and dynamics than our model. Our model uses some maps and parameter values from the PTBS reference model, and the other model parameters are hand-tuned to reduce errors between the two models.

After calibration, we validate our model by testing and comparing the $MPGe$ of our model and that of the PTBS reference model for different driving cycles. The $MPGe$ mismatches between the two models for the Urban Dynamometer Driving Schedule (UDDS), the Highway Fuel Economy Test (HWFET), and the US06 driving cycles are 5.94%, 5.90%, and 7.95%, respectively. Fig. 1.2 shows the time histories of powertrain signals for the BEV driving through the UDDS cycle, where the blue curves correspond to our model after calibration and the red curves correspond to the PTBS reference model. It can be observed that the signals of our model closely match those of the PTBS reference model. These results validate that our model (1.1)-(1.16) after calibration can be used to produce sufficiently accurate energy consumption estimates (accurate in terms of matching the estimates produced by the high-fidelity PTBS reference model). Note that our model entails much lower computational footprint than the PTBS reference model, and is thus suitable for RL purposes.

The BEV powertrain model (1.1)-(1.16) is then converted to discrete-time, assuming a 1-second sampling period, and integrated with the traffic simulator of [15, 16], used for the RL-based development of energy-efficient autonomous vehicle control policy.

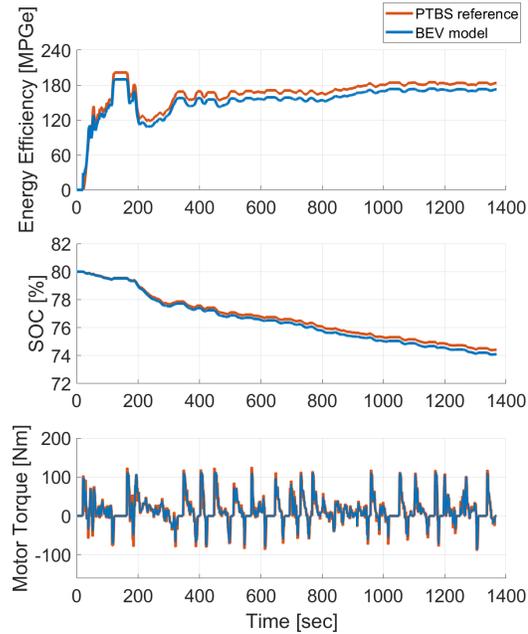

**Fig. 1.2** Time histories of powertrain signals for the UDDS cycle.

## 1.4 Controller Design

### 1.4.1 Game-Theoretic Traffic Environment

In order to train an autonomous vehicle control policy offline, we use an interactive traffic simulator, following the approach described in [15], as the training environment. This level-$k$ game-theory based simulator assumes that the traffic consists of human-driven vehicles that can be modeled by cognitive behavioral models with $k$ levels. Studies such as [18] show that human reasoning process rarely exceeds three steps, so $k = 0, 1, 2$ is used in this work. The level-0 vehicles use a hand-crafted policy that commands one of the three actions, "maintain speed", "decelerate", or "hard decelerate", based on the range and range rate with the vehicle in its front to represent vehicle behavior under minimal rationality; level-1 and level-2 vehicles use policies trained by RL assuming that the surrounding vehicles are all level-0 and level-1, respectively. Moreover, the level-1 and 2 vehicles have a larger action space consisting of seven actions, including acceleration, deceleration, and lane change. Overall, the results of [15] indicate that the level-0 drivers/vehicles have the most conservative behaviors, while level-1 vehicles behave the most aggressively such as driving faster and frequently making lane changes, and the aggressiveness of level-2 vehicles falls between level-0 and level-1.

*Remark 2* Similarities and differences between the setup considered in this work and that in [15] are highlighted in Table 1.2, in terms of models, reward functions, surrounding traffic, observation space and action space, RL algorithm, and training process. Details are given in the subsequent subsections.

**Table 1.2** Comparison summary on control development for the level-$k$ ($k = 1, 2$) policies and the autonomous vehicle policy considered in this work (AV)

|  | Level-$k$ for $k = 1, 2$ | AV |
|---|---|---|
| BEV model | Not included | Included |
| Reward function | $R_1$ | $R_1 + R_2$ |
| Surrounding traffic during RL | Level-$(k-1)$ | A mixture of level-0, level-1, and level-2 with a certain ratio |
| Observation space | 11 states | 11 states plus $V$ and $SOC$ |
| Action space | 7 actions | Same as level-$k$ |
| RL algorithm | Jaakkola RL algorithm | Same as level-$k$ |
| After training, assign level-0 policy to states visited fewer than $n$ times | $n = 20$ | $n = 40$ |

### 1.4.2 Observation and Action Spaces

The observation space, i.e., input space, for our autonomous vehicle control policy is extended from the observation space for the level-$k$ driver policies. The observation space for the level-$k$ policies has 11 observations, including the range and range rate of the ego vehicle to the vehicles in its front, front right, front left, rear right, and rear left, as well as the lane index of the ego vehicle. The range is categorized by "far", "nominal", or "close", and the range rate is categorized by "moving away", "stable", and "approaching". The simulator is configured for a three-lane highway. As a result, the total number of possible states in the level-$k$ policy is $3^{11}$. Here, a state means a unique combination of observations.

To incorporate considerations of energy efficiency, it is necessary to enlarge the observation space by including additional observations. Since vehicle speed and battery $SOC$ are critical factors that affect the PCU decision for the regenerative power distribution, as well as the component efficiencies of the battery and the MG, they are added to the observation space, each being categorized by "high", "medium", and "low". Consequently, the total number of possible states for our autonomous vehicle control policy increases to $3^{13}$.

The action space, i.e., output space, for the proposed control design is the same as that for the level-$k$ driver policies. It includes the following seven actions: 1) maintain speed, 2) accelerate, 3) decelerate, 4) hard accelerate, 5) hard decelerate, 6) move left (if the vehicle is not in the left-most lane) and 7) move right (if the vehicle is not in the right-most lane).

The exact definitions of the observation and action spaces for the level-$k$ driver policies are given based on the parameters including the relative longitudinal position and speed thresholds, acceleration rates, deceleration rates, and lane change rates, whose values are the same as in [15], so they are not repeated here. For the two additional observations of the autonomous vehicle control policy, i.e., the vehicle speed and battery $SOC$, the thresholds that divide the three categories are, respectively, 17.22 m/s and 22.22 m/s, and 70% and 80%, chosen by trial and error.

### 1.4.3 Reward Function

The reward function used for RL training is as follows,

$$\mathcal{R} = R_1 + R_2, \tag{1.17}$$

where

$$R_1 = w_1 \cdot c + w_2 \cdot v + w_3 \cdot h + w_4 \cdot u, \tag{1.18}$$

$$R_2 = w_5 \cdot e. \tag{1.19}$$

1 Energy-Efficient Autonomous Driving Using Reinforcement Learning 11Here, $w_i > 0$, $i = 1, \ldots, 5$ are weights, and $c$, $v$, $h$, $u$, and $e$ are reward features. In particular, the reward function consists of two parts: The first part $R_1$ with the terms $c$, $v$, $h$, and $u$ accounts for the safety, performance, and comfort requirements and shares the same setup as for the level-$k$ policies. The second part $R_2$ is an additional term that accounts for energy efficiency.

The reward features and their corresponding weights are chosen based on engineering insight and tuning by simulation as follows:

- $c$ accounts for constraint violations,

$$c = \begin{cases} -1, & \text{if a collision occurs to the ego vehicle,} \\ 0, & \text{otherwise,} \end{cases} \quad (1.20)$$

  with $w_1 = 10,000$.

- $v$ accounts for travel speed,

$$v = \frac{V - v_n}{a}, \quad (1.21)$$

  where $V$ is the speed of the ego vehicle in the longitudinal direction, and the constants $v_n$, a nominal speed, and $a$, a nominal acceleration rate, are used to scale this term to the same order of magnitude of the other terms, with $w_2 = 5$.

- $h$ accounts for headway, encouraging the ego vehicle to keep a reasonable distance from preceding vehicles,

$$h = \begin{cases} 1, & \text{if headway} \in \text{"far",} \\ 0, & \text{if headway} \in \text{"nominal",} \\ -1, & \text{if headway} \in \text{"close",} \end{cases} \quad (1.22)$$

  with $w_3 = 1$. Here, "headway" means the range of the ego vehicle to the vehicle in its immediate front.

- $u$ accounts for control effort,

$$u = \begin{cases} 0, & \text{if action = "maintain speed",} \\ -1, & \text{if action = "accelerate" or "decelerate",} \\ -3, & \text{if action = "move left" or "move right",} \\ -5, & \text{if action = "hard accelerate" or "hard decelerate",} \end{cases} \quad (1.23)$$

  with $w_4 = 1$.

- $e$ is for energy efficiency, defined by the time derivative of $MPGe$ as

$$e = \frac{\mathrm{d}MPGe}{\mathrm{d}t} = \frac{V \cdot E_{batt} - x \cdot P_{batt}}{E_{batt}^2} \cdot \gamma, \quad (1.24)$$

  with $w_5 = 5$.

### 1.4.4 Training Algorithm

The goal of training is to find a control policy that maximizes the reward averaged over an infinite horizon. Using the settings described above, we formulate this problem as a partially observable Markov decision process (POMDP) problem since only certain observations are available to the ego vehicle. For example, if there are multiple vehicles in front of the ego vehicle on the same lane, the ego vehicle can only observe the relative range and range rate of the vehicle immediately in front of it. Thus, the algorithm used for training the control policy should guarantee convergence of the average reward with respect to POMDP problems. We choose to use the Jaakkola RL algorithm [49] since, under suitable assumptions, this algorithm guarantees convergence of the average reward to a local maximum for POMDP problems. The proof of such a convergence guarantee can be found in Appendix A of [50].

A summary of the Jaakkola RL algorithm is given in [15] and Section 1.2.7 of [50]. The key variables and equations are reviewed here. The algorithm iterates with two steps at every simulation time step $t$. First, the one-step reward $R_t$ is evaluated based on the results of the simulation following the current policy $\pi_t$. Then, for each observation state $o \in O$, and state and action pair $(o,a) \in O \times \mathcal{A}$, the state-value functions $V(o|\pi_t)$ and the action-value functions $Q(o,a|\pi_t)$, also called Q-values, are updated based on the difference of $R_t - \bar{R}(\pi_t)$ where $\bar{R}(\pi_t)$ is the average reward for an infinite duration with the policy $\pi_t$. The state-value $V(o|\pi_t)$ represents the expected cumulative reward starting at state $o$ following policy $\pi_t$, while the Q-values $Q(o,a|\pi_t)$ represents the expected cumulative reward if the state starts at $o$, we take action $a$ first, and then follow policy $\pi_t$ afterward. Specifically, the state-value functions and Q-values are updated with equations given as

$$\beta_t^o(o) = \left(1 - \frac{\chi_t^o(o)}{K_t^o(o)}\right)\gamma_t \beta_{t-1}^o(o) + \frac{\chi_t^o(o)}{K_t^o(o)}, \tag{1.25}$$

$$V(o|\pi_t) = \left(1 - \frac{\chi_t^o(o)}{K_t^o(o)}\right) V(o|\pi_{t-1}) + \beta_t^o(o)\left(R_t - \bar{R}(\pi_t)\right), \tag{1.26}$$

$$\beta_t^a(o,a) = \left(1 - \frac{\chi_t^a(o,a)}{K_t^a(o,a)}\right)\gamma_t \beta_{t-1}^a(o,a) + \frac{\chi_t^a(o,a)}{K_t^a(o,a)}, \tag{1.27}$$

$$Q(o,a|\pi_t) = \left(1 - \frac{\chi_t^a(o,a)}{K_t^a(o,a)}\right) Q(o,a|\pi_{t-1}) + \beta_t^a(o,a)\left(R_t - \bar{R}(\pi_t)\right), \tag{1.28}$$

where $\chi$ represents a binary (0 or 1) indicator function that equals to one if $o$ or $(o,a)$ is visited at the current time step, $K$ is a function that counts how many times $o$ or $(o,a)$ has been visited, and $\gamma_t$ is a time-dependent discount factor that takes a value between zero and one and converges to one as $t$ goes to infinity. In the second step, the policy is updated by the following equation

$$\pi_{t+1}(o,a) = (1-\epsilon)\pi_t(o,a) + \epsilon\hat{\pi}_t(o,a), \quad \forall (o,a) \in O \times \mathcal{A}, \qquad (1.29)$$

where $\epsilon \in (0,1)$ is the learning rate and $\hat{\pi}_t$ is the greedy policy that maximizes

$$J_t(\pi,o) = \sum_{a \in \mathcal{A}} \pi(o,a)\Big(Q(o,a|\pi_t) - V(o|\pi_t)\Big), \quad \forall o \in O. \qquad (1.30)$$

The process then moves on to the next time step and the iteration of the above two steps continues.

Note that the Jaakkola RL algorithm updates the value functions and Q-values at each time step using both the immediate one-step reward $R_t$ and the infinite-horizon average reward $\bar{R}$. The one-step reward can make the policy update respond instantly to the large penalty of a safety constraint violation. The average reward, which in actual implementation is estimated by averaging all past instant rewards, eventually contains the weighted energy efficiency $MPGe$ computed over a long horizon. In this way, objectives with different time horizons are handled simultaneously.

### 1.4.5 Training Process

The level-$k$ policies with $k = 1$ and 2 are obtained following Algorithms 1 and 2 of [15] (similar to Algorithm 1 below) with the reward function $R_1$ described above, and thus, they do not account for energy efficiency. We then use vehicles operating with level-$k$ policies to provide the traffic environment for the RL training of our autonomous vehicle controller that considers energy efficiency.

The training process for the proposed autonomous vehicle control policy is summarized by the pseudo-code in Algorithm 1. Each training episode corresponds to a simulation trajectory with a duration of 200 seconds. This RL training process is similar to the process described by Algorithms 1 and 2 of [15] with the following major differences: 1) When initializing a training episode, we initialize the ego vehicle battery $SOC$ randomly according to a uniform distribution on the range of $[15\%, 90\%]$. For Algorithm 2 of [15], however, since $SOC$ is not a state of the level-$k$ models, this initialization step does not exist; 2) A traffic environment consisting of a mixture of level-0, 1, and 2 vehicles with a ratio of 15%, 55%, and 30% is used to train our autonomous vehicle control policy, while when training a level-$k$ policy, by definition, vehicles in the environment are all level-$(k-1)$; 3) Due to the increase of the size of the observation space, the total number of possible observation combinations is 9 times greater than for the level-$k$ policies. With the same number of training episodes (i.e., 50,000, determined/limited by the affordable computational resources such as training time duration), it is more likely that some states are not sufficiently visited during the training. Thus, an increased value of the parameter $n$ is used in the last step for the autonomous vehicle policy training, where $n$ represents the number of times a state has to be visited during training, lest it be assigned the level-0 policy.

**Algorithm 1:** Training process

1. Initialize the ego car's policy with equal action probabilities for every state.
2. $episode \leftarrow 0$.
3. **while** $episode \leq 50,000$ **do**
4.     Randomly select the number of surrounding cars, $n_c \in [21, 30]$.
5.     Initialize surrounding cars with level-$k$ policies with probabilities of 15%, 55% and 30% for $k = 0, 1$ and 2.
6.     Initialize the ego car with $SOC \in [15\%, 90\%]$.
7.     **while** $t \leq 200$ **do**
8.         Run simulation and evaluate the ego car's policy with the reward function $\mathcal{R}$.
9.         Update the ego car's policy.
10.         **if** *a collision occurs to the ego vehicle* **then**
11.             Terminate the current episode.
12.         **end if**
13.     **end while**
14.     $episode \leftarrow episode + 1$.
15.     Assign the level-0 policy to states visited less than $n = 40$ times.
16. **end while**

### 1.4.6 Autonomous Vehicle Control Policy for Benchmarking

For comparison purposes, a second RL-based policy is trained in the mixed traffic environment described above. This benchmark policy uses only $R_1$, as defined in (1.18), as its reward function, and allows one to study the differences between considering the fuel economy or not, in similar traffic conditions.

In addition to policies trained by RL, the FSM-based policy described in [16] is adopted for comparison. The FSM-based policy is a rule-based controller with three modes including cruise control, adaptive cruise control, and lane change control. Switches between modes are triggered when certain traffic conditions are satisfied. The FSM-based policy is calibrated to optimize safety and performance, while the energy efficiency is not being considered.

## 1.5 Results

### 1.5.1 Training for RL-Based Policies

As discussed above, each RL-based policy is trained for 50,000 episodes. Fig. 1.3 shows the values of the average reward as the training progresses for the level-1 policy, level-2 policy, the proposed autonomous vehicle control policy with the energy efficiency consideration (AV w/ $e$), and the benchmark policy that does not account for the energy efficiency (AV w/o $e$). It can be observed that the average rewards all converge smoothly, suggesting the success of the RL procedures.

The value of the converged average reward of each policy is a combined result of the different reward features in $\mathcal{R}$. For example, the converged average reward of the level-2 policy is higher than those of the other policies. This is because, among the level-0, 1, and 2 policies, the level-1 policy is the most aggressive as concluded in [15], tending to make many lane changes to pursue higher travel speeds. Since the traffic environment for training the level-2 policy is composed of purely level-1 vehicles, the level-2 policy is relatively conservative and collisions are less likely. Moreover, the overall faster traffic flow allows a higher travel speed of the ego vehicle. Thus, the coupled effect of these factors leads to a higher converged average reward for the level-2 policy.

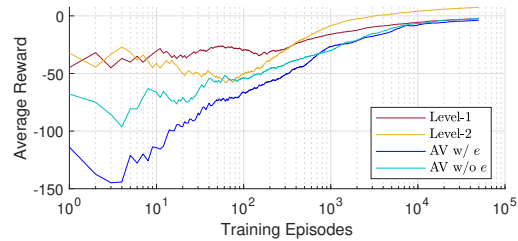

**Fig. 1.3** Average reward evolution during RL.

## 1.5.2 Control Performance

### 1.5.2.1 Evaluation Process

The control policies are evaluated based on simulations using the process described by the pseudo-code in Algorithm 2. In particular, for each policy and each traffic density (represented by the number of surrounding vehicles in traffic, ranging from 0 to 30), 10,000 simulation episodes are run, each with a duration of 200 seconds. Then, the policy is evaluated with respect to four evaluation metrics, including:

- Constraint violation rate, defined as the percentage of simulation episodes where a collision occurs to the ego vehicle;
- Average number of lane changes per simulation episode;
- Average $MPGe$;
- Average travel speed.

**Algorithm 2:** Evaluation process

1  **for** $n_c = 0 : 30$ **do**
2      $episode \leftarrow 0$.
3      **while** $episode \leq 10,000$ **do**
4        Initialize the ego car with $SOC \in [15\%, 90\%]$ and the control policy to be evaluated.
5        Initialize surrounding cars with level-$k$ policies randomly with probabilities of 15%, 55% and 30% for $k = 0$, 1 and 2.
6        Simulate and record variables relevant to the evaluation metrics.
7        $episode \leftarrow episode + 1$.
8      **end while**
9      Compute and output the evaluation metric values.
10 **end for**

#### 1.5.2.2 Simulation Result Analysis

*Proposed AV control policy*

Figs. 1.4 to 1.6 show the results of different policies in regards to the four aforementioned evaluation metrics as functions of the traffic density. Figs. 1.4(a) and 1.5 show the constraint violation rate, and Figs. 1.4(b) and 1.6(a) show the average number of lane changes. It can be observed that, when driving in the mixed traffic environment (vs. Mix), the proposed policy (AV w/ $e$) has the lowest constraint violation rate among all policies that can perform lane changes, indicating good safety features.

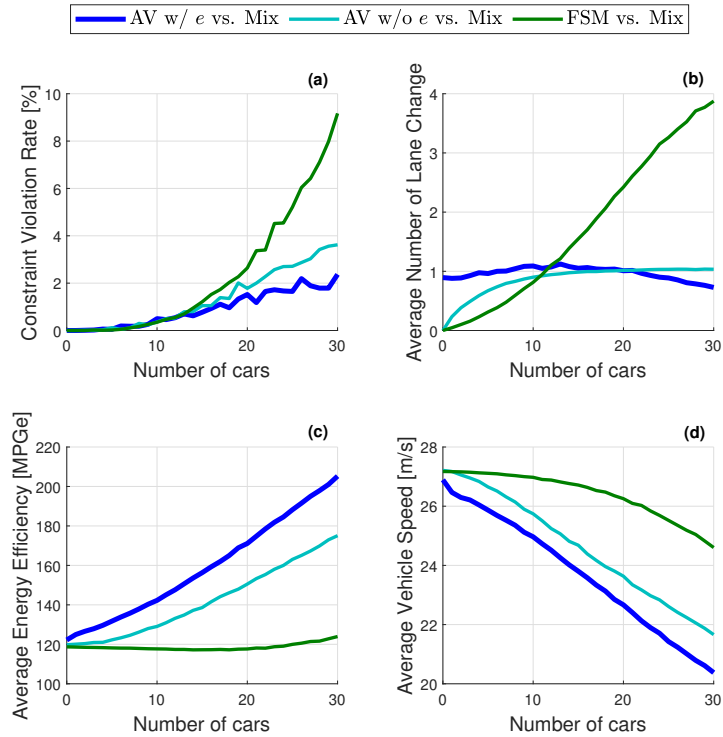

**Fig. 1.4** Evaluation results for AV control policies in traffic environments of different traffic densities: (a) Constraint violation rate, (b) Average number of lane changes per simulation episode, (c) Average $MPGe$, (d) Average travel speed.

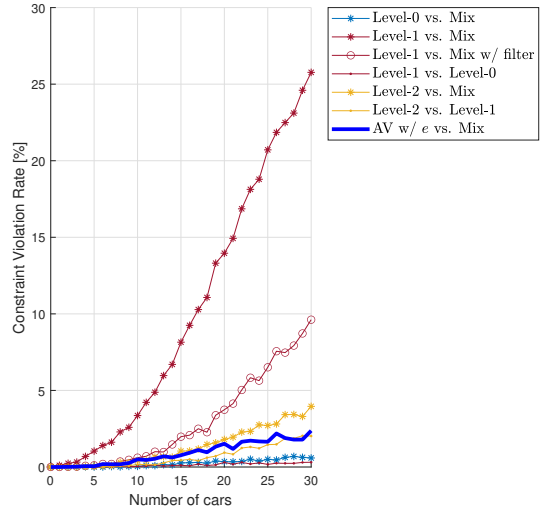

**Fig. 1.5** Constraint violation rates for level-$k$ and the proposed AV control policies in traffic environments of different traffic densities.

Fig. 1.4(c) compares the average $MPGe$ among the three AV control policies, i.e., the proposed policy (AV w/ $e$), the RL-based benchmark policy (AV w/o $e$), and the FSM-based policy. It shows that the proposed policy with the fuel economy consideration is more energy-efficient than the other two policies, verifying that the additional observations (vehicle speed and $SOC$) and the fuel economy term $R_2$ in the reward function $\mathcal{R}$ are effective for the improvement of energy efficiency.

The average travel speed of each policy is shown in Figs. 1.4(d) and 1.6(b). It can be observed that at low traffic density, the autonomous vehicle controlled by the proposed AV policy drives at a higher average speed, close to that of level-2 vehicles; but when the traffic gets denser, the autonomous vehicle slows down to an average speed close to that of a level-0 vehicle. This can be explained with the help of Fig. 1.4(b): The average number of lane changes of the proposed policy varies only slightly for different traffic densities, since when the traffic density is low, there is not much need to change lanes, while when the traffic density gets high, it may be neither safe nor energy-efficient to perform lane changes. This feature distinguishes the behavior of the proposed policy from the level-1 policy and the FSM-based policy that prefer higher travel speeds and thus make many lane changes to achieve them.

*Miscellaneous Observations*

The results indicate that the effects of the different evaluation aspects, that are closely related to the five features in the reward function, are not decoupled and can largely affect each other. For example, for the level-$k$ policies, although the level-0 policy has the lowest constraint violation rate, it is a very conservative policy that does not allow lane changes (as illustrated in Fig. 1.6(a)) and has the lowest average vehicle speed (as shown in Fig. 1.6(b)). When driving in the level-0 environment (vs. Level-0), consequently, the level-1 policy also has a very low constraint violation rate. However, when driving in the mixed traffic environment, where there are other level-1 cars and level-2 cars, the level-1 policy has the highest constraint violation rate instead.

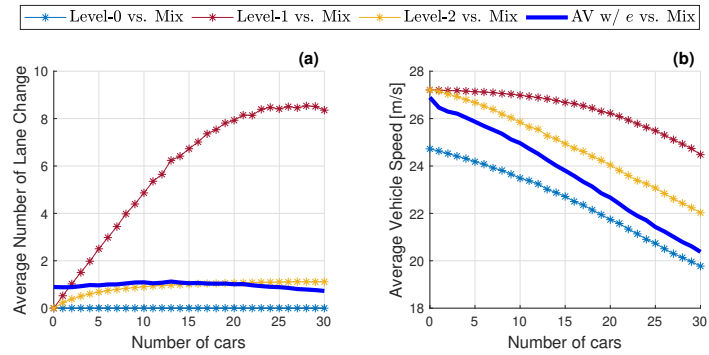

**Fig. 1.6** Average number of lane changes per simulation episode and average travel speed for level-$k$ and the proposed AV control policies in traffic environments of different traffic densities.

It is worth highlighting some additional observations in the simulation results. Firstly, for policies trained by RL, the average $MPGe$ increases as the average travel speed decreases in denser traffic. This is attributed to the fact that for highway driving, the energy efficiency at the vehicle level is affected largely by energy losses from rolling and aerodynamic resistance that increase as the vehicle travel speed increases. To see the significant impact of rolling and aerodynamic resistance on energy consumption, let us consider and compare two cases: In the first case, the vehicle is driving at a constant speed of 20.5 $m/s$. In the second case, the vehicle is driving at 24.5 $m/s$. These two cases roughly represent, respectively, the average longitudinal behavior of the proposed AV policy and that of the FSM-based policy with 30 surrounding vehicles, shown in Fig. 1.4(d). Here, we ignore the differences in the MG efficiency by assuming a constant value $\eta_0$. Then, we have that the MG power consumed by the rolling and aerodynamic resistance can be calculated as

$$P_l = (T_l \cdot \frac{1}{g}) \omega_{mg}/\eta_0$$
$$= (a + b \cdot V + c \cdot V^2)V / (r\eta_0), \qquad (1.31)$$

depending cubically on the vehicle speed $V$. Without considering the discrepancy in the battery and MG efficiency, we then use (1.31) to estimate the difference in the MG power used to counteract the rolling and aerodynamic resistance. We obtain that the discrepancy between the two cases is about 33%. This contributes to the difference in the $MPGe$ shown in Fig. 1.4(b), where the change is about 64%. It is within a reasonable range according to Table 1.1 (e.g., [11, 13]).

Note, however, that the proposed policy does not always operate the autonomous vehicle at a low speed, as the reward function represents several different objectives besides energy efficiency. In general, energy efficiency depends on the traffic scenario, the powertrain type, as well as the component specifications. This fact highlights the benefit of modeling the powertrain system using maps, so that components can be easily sized or swapped.

Secondly, it is observed in Fig. 1.4(b) that when there is no other vehicle in traffic ("zero-traffic"), the autonomous vehicle controlled by the proposed policy makes on average one lane change. This is because the policy by RL converges to a solution where when there is no other vehicle in the immediate vicinity of the ego vehicle, the ego vehicle tends to change to and stay in the right-most lane to reduce the possibility of having interactions/conflicts with other vehicles that may degrade its safety and energy efficiency in the future. Note also that such a solution may only be locally optimal (i.e., not globally optimal), as the Jaakkola RL algorithm used to train the policy guarantees only convergence to a local optimum (and not necessarily the global optimum) [49].

Intuitively speaking, vehicles need not change lanes when there are no slower vehicles in their front blocking their ways. This is the case for most policies shown in Figs. 1.4(b) and 1.6(a) except for the proposed policy for which the average number of lane changes in the zero-traffic environment is close to but slightly less than one. For most cases, when initialized in the middle lane, the autonomous vehicle controlled by the proposed policy immediately makes a lane change to the right, as explained above. However, for some states with high SOC that were not visited enough during RL training, the policy was overridden by the level-0 policy that does not perform lane changes (see the last line of Algorithm 1). This caused the average number of lane changes to be slightly less than one. Note also that the training is conducted only for traffic environments with 21 to 30 surrounding vehicles, i.e., not covering the zero-traffic environment. Such sub-optimal behavior in the zero-traffic environment of the trained policy indicates that it might be beneficial to conduct training for a wider range of traffic environments if computational resources allow.

Thirdly, one of the major contributors to the high constraint violation rate of the level-1 policy when operating in the mixed traffic environment is its high frequency of lane changes. Two problematic scenarios related to lane changes have been identified in [15]. The first case involves a scenario where the ego vehicle originally driving in the right (or left) lane and another vehicle originally driving in the left (or right) lane in an almost parallel longitudinal position with the ego vehicle simultaneously start to perform lane changes to the middle and lead to a side collision between them. The second case involves a scenario where the ego vehicle starts to change lanes trying to overtake a vehicle in its front, but at the same time, the preceding vehicle also starts to change lanes in the same direction (e.g., trying to overtake another vehicle) and blocks the ego vehicle's overtaking. Since the level-1 policy is trained using an environment consisting of only level-0 vehicles that do not change lanes, these two "unrare in reality" scenarios have never occurred during the RL training. Consequently, the level-1 policy fails to learn to avoid such scenarios. We have identified all constraint violation cases in the Level-1 vs. Mix data that belong to these two scenarios and computed the constraint violation rate after filtering out these cases. The result is plotted in Fig. 1.5, called Level-1 vs. Mix w/ filter. It can be seen that the constraint violation rate of the level-1 policy after this filtering is significantly reduced.

If such an issue happens when developing autonomous vehicle control algorithms, where problematic scenarios can be clearly identified, they can be handled by specific add-on mechanisms. For example, the autonomous vehicle may be commanded to go back to its original lane when either of the above two cases is detected.

## 1.6 Conclusions

In this chapter, an autonomous vehicle control policy is developed focusing on energy efficiency optimization while safety, performance, and comfort are balanced. We first discuss the potential of autonomous vehicle (AV) controls for energy-efficient driving and the major challenges to develop such control policies. Then, we show the powertrain model built to capture the energy consumption of a battery electric vehicle (BEV), integrated with the highway traffic simulator consisting of cognitive driver behavioral models based on level-$k$ game theory. An AV control policy is trained by reinforcement learning (RL) for this BEV and compared with two benchmark policies as well as the level-$k$ policies from different evaluation perspectives.

Analysis of the results indicates that the addition of the energy efficiency term in the RL reward function, in addition to the expanded observation space to include the vehicle speed and $SOC$, is effective in improving the energy efficiency while maintaining low collision rates. Through analysis of the BEV powertrain model, the increase of the energy efficiency represented by $MPGe$ is likely dominated by the reduction of the average vehicle speed that lowers the rolling and aerodynamic resistance. However, this does not make the vehicle always travel at the lower speed limit, which highlights the capability of the RL-based approach that does a good job in balancing travel speeds, safety, and efficiency. The results also imply the potential to further extend and explore the control design in terms of higher computational efficiency and advanced RL algorithms for control performance improvement. In the future, our AV policy may serve as a baseline control strategy for more advanced autonomous driving control development.